\documentclass[runningheads]{llncs}
\usepackage[T1]{fontenc}
\usepackage[utf8]{inputenc}
\usepackage{amsmath,amssymb,amsfonts,bm}
\usepackage{graphicx,tabularx}
\usepackage{microtype}
\usepackage{lmodern}
\usepackage[english]{babel}
\usepackage[bottom]{footmisc}
\usepackage{booktabs}
\usepackage{hyperref}
\hypersetup{
    colorlinks=true,
    linkcolor=black,
    filecolor=black,
    citecolor=blue,
    urlcolor=black,
}

\begin{document}

\title{
    Learning and Clustering on Temporal Graphs:\\Principles, Primitives, and Pooling
}
\titlerunning{Learning and Clustering on Temporal Graphs}

\author{
    Nelson Aloysio Reis de Almeida Passos\inst{1,2}\orcidID{0000-0003-1869-2976} \and
    Emanuele Carlini\inst{2}\orcidID{0000-0003-3643-5404} \and
    Salvatore Trani\inst{2}\orcidID{0000-0001-6541-9409}
}
\authorrunning{N. A. R. A. Passos et al.}

\institute{
    University of Pisa, Dept. of Computer Science, 56127 Pisa PI, Italy \and
    National Research Council, 56124 Pisa PI, Italy \\[1em]
    \email{
        nelson.reis@phd.unipi.it \\
        \{emanuele.carlini,salvatore.trani\}@isti.cnr.it}}

\maketitle

This work\footnote{
    Accepted at ECML PKDD 2026 (Nectar Track), based on prior work \cite{passos2024_gnnet,passos2025_tadcsbm,passos2026_frame,passos2026_symposium}.
} focuses on the problem of learning on temporal graphs, with a particular emphasis on the principles and primitives surrounding the task of \textit{clustering} \cite{dmon_tsitsulin2024}: obtaining coarse-grained representations by aggregating information from nodes, edges, and temporal dynamics --- a task related to that of \textit{pooling} \cite{pooling_grattarola2021} in machine learning on graphs, or \textit{community detection} \cite{peixoto2023} in network science.

We depart from a question intersecting both literatures: whether
learning on graphs and community detection reinforce one another, compete, or prove largely orthogonal in practice.
Graph neural networks (GNNs) have reached state of the art for diverse downstream tasks, yet their advantage over established descriptive and inferential clustering algorithms \cite{peixoto2023} remains far less settled \cite{survey_longa2023}, particularly under joint demands of computational efficiency and recovery accuracy \cite{peel2017_groundtruths}.

Our central claim is that temporal clustering becomes relevant to graph learning when treated simultaneously as a statistical \textit{principle}, a scalable computational \textit{primitive}, and a \textit{pooling} mechanism.
An initial study of ours found that augmenting neural models with temporal structure did not consistently improve the recovery of ground-truth communities over algorithmic baselines \cite{passos2024_gnnet}; a finding later corroborated on synthetic graphs sampled, under known ground truth, from a time-varying, attributed, degree-corrected stochastic block model we designed for this purpose \cite{passos2025_tadcsbm}.
For non-attributed temporal community detection, principled algorithmic methods remain the appropriate tool, and the decisive obstacle is scalability rather than accuracy; the case for neural models is strongest precisely in the attributed regime that descriptive spectral and modularity \cite{spectral_modularity_newman2006} methods do not exploit in full, and even there the gains are clearest where structural, attribute, and temporal signals align, rather than as a universal advantage \cite{peixoto2023}.

\smallskip\textbf{Principles.} Community detection rests on the premise that nodes organize into groups whose connectivity --- and, where present, attributes and temporal behavior --- set them apart from the rest of the network \cite{peixoto2023}.
The prevailing notion is assortative: groups more densely connected within than without, the definition underpinning descriptive quality functions such as modularity and relaxations solved on graph spectra \cite{spectral_modularity_newman2006}.
A sharper, theory-grounded criterion is the \textit{detectability} of communities: in the sparse stochastic block model (SBM) there exists a threshold --- the Kesten--Stigum bound $c\lambda^2 = 1$, with $\lambda$ the community signal and $c$ the average degree --- below which no efficient algorithm recovers planted structure better than chance in the large-graph limit \cite{detectability_ghasemian2016}.
Ordinary Laplacian spectra fall short of this bound, their informative eigenvectors localizing on high-degree nodes; operators built on the non-backtracking (Hashimoto) matrix instead remain informative down to the threshold itself \cite{bethehessian_saade2014}.
The same theory extends to the temporal domain, where a second parameter $\eta \in [0,1]$ --- the rate at which nodes retain their community label across snapshots --- couples spatial and temporal signal, so that detectability comes to depend jointly on $\lambda$ and $\eta$ \cite{detectability_ghasemian2016}.

Crucially, spectral theory also underlies graph learning.
A node signal $\mathbf{x}$ is projected to the spectral domain as $\hat{\mathbf{x}} = \mathbf{U}^\top \mathbf{x}$, with $\mathbf{U}$ the Laplacian eigenbasis, so that a learned filter acts as $\mathbf{y} = \mathbf{U} g_\theta(\mathbf{\Lambda}) \mathbf{U}^\top \mathbf{x}$; graph convolutions are localized polynomial approximations of this construction, aggregating over $K$-hop neighborhoods without explicit eigendecomposition \cite{sgcn_kipf2017}.
So community detection and graph learning draw on a common foundation: the operators that reveal mesoscale structure are those on which message passing is defined.
It is this principled footing that motivates spectral and modularity-based methods as the substrate for end-to-end differentiable clustering \cite{dmon_tsitsulin2024} --- a choice that, as we show, has practical consequences for a GNN's performance, not merely theoretical ones.

\smallskip\textbf{Primitives.} Algorithmic methods are traditionally CPU-bound and ill-suited to large temporal graphs: temporal coupling of nodes through time enlarges the adjacency matrix to order $nT \times nT$, so cost grows prohibitively with both the number of nodes and the number of time steps \cite{modularity_mucha2010}.
Our recent contribution makes them tractable at scale --- GPU-accelerated multislice modularity optimization \cite{modularity_mucha2010,leiden_traag2019} and spectral formulations \cite{spectral_modularity_newman2006,bethehessian_saade2014} over sparse supra-adjacency representations, built on the CuPy and RAPIDS (cuGraph, cuML) ecosystem \cite{passos2026_frame}.
However, the spectral case is non-trivial on GPUs: accelerated sparse eigensolvers are currently restricted to symmetric (real) or Hermitian (complex) operators, whereas the non-backtracking matrix attaining the detectability threshold on temporal graphs is, by construction, asymmetric --- aptly, as information flows forward in time.

\begin{figure*}[b!]
\vspace{-1em}
    \centering
    \begin{minipage}[t]{0.52\textwidth}
        \vspace{-0.7em}
        \centering
        \includegraphics[width=\linewidth]{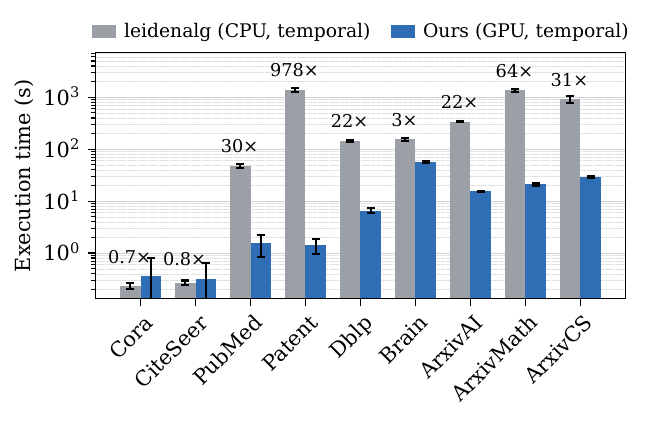}
    \end{minipage}%
    \hfill
    \begin{minipage}[t]{0.48\textwidth}
        \vspace{0pt}
        \centering
        \scriptsize
        \renewcommand{\arraystretch}{1.27}
        \setlength{\tabcolsep}{2.7pt}
        \resizebox{\linewidth}{!}{%
        \begin{tabular}{lrrrrr}
            \toprule
            \textbf{Dataset} & $\boldsymbol{|\mathcal{V}|}$ & $\boldsymbol{\Sigma(\mathcal{V})}$ & $\boldsymbol{|\mathcal{E}|}$ & $\boldsymbol{\Sigma(\mathcal{E})}$ & $\boldsymbol{|T|}$ \\
            \midrule
            Cora~$\dagger$ & 2\,708 & 2\,708 & 5\,278 & 5\,278 & 1 \\
            CiteSeer~$\dagger$ & 3\,279 & 3\,279 & 4\,552 & 4\,552 & 1 \\
            \midrule
            PubMed & 19\,717 & 37\,003 & 44\,324 & 44\,324 & 42 \\
            Patent & 12\,214 & 41\,529 & 41\,916 & 41\,916 & 891 \\
            Dblp & 28\,085 & 101\,797 & 153\,822 & 222\,165 & 27 \\
            Brain & 5\,000 & 60\,000 & 878\,207 & 947\,744 & 12 \\
            \midrule
            ArxivAI & 69\,854 & 142\,316 & 696\,819 & 696\,819 & 27 \\
            ArxivMath & 270\,013 & 614\,578 & 783\,165 & 783\,165 & 31 \\
            ArxivCS & 169\,343 & 374\,433 & 1\,157\,799 & 1\,157\,799 & 29 \\
            \bottomrule
        \end{tabular}%
        }
    \end{minipage}
    \vspace{-0.6em}
    \caption{
        \footnotesize
        \textbf{Runtime comparison} \cite{passos2026_frame}.
        Left: CPU/GPU runtimes over five runs, mean $\pm$ standard deviation (log scale).
        Right: Datasets \cite{data4tgc2023,passos2024_gnnet}: $|\mathcal{V}|,|\mathcal{E}|$ are unique nodes/edges, $\Sigma(\mathcal{V}),\Sigma(\mathcal{E})$ are sums over snapshots, and $|T|$ is the snapshot count; $\dagger$ marks static baselines.
        Matched two-iteration Leiden budget (\texttt{leidenalg} default) on an Intel Xeon Gold 6330 (CPU) or NVIDIA A100 80GB (GPU), including host-to-device transfer.
    }
    \label{fig:results}
\end{figure*}

We escape the asymmetry by reformulating the eigenproblem: solving the symmetric Bethe-Hessian $\mathbf{H}(r) = (r^2 - 1)\mathbf{I} - r\mathbf{A} + \mathbf{D}$, whose negative eigenvalues recover the informative, out-of-bulk directions of the non-backtracking operator \cite{bethehessian_saade2014} --- preserving asymptotic optimality for undirected graphs while keeping matrix operations fully on GPU.
In static SBM graphs, $r \approx \sqrt{c}$, while the temporal case requires coupling $\lambda$ and $\eta$ \cite{detectability_ghasemian2016} --- hence a different regularization, as likely does the attributed regime.
Against a CPU reference under an equal-work budget, our multislice modularity backend attains speedups of up to roughly three orders of magnitude, depending on graph density and snapshot count (Figure~\ref{fig:results}).
For the largest graphs, the effect is a change in tractability: temporal graphs that exceed practical time limits on CPU become routine on GPU.
Exposed as a zero-code-change backend of the NetworkX-Temporal library \cite{networkxtemporal2025}, the implementation requires only a single environment variable to move a pipeline from CPU to GPU.

\smallskip\textbf{Pooling.} This primitive is of interest beyond community detection itself.
Learning on very large graphs is increasingly limited by scale, and a standard response is coarse-graining: reducing the graph to a smaller representation on which a model can operate, then refining or unpooling the result \cite{pooling_grattarola2021}.
This reduction is rarely inconsequential --- most graph pooling is heuristic or learned without guarantees --- and a principled community detector offers an alternative: a coarse-graining operator grounded in detectability theory, where the temporal supra-graph assigns labels jointly across snapshots, dispensing with post-hoc matching between independently detected partitions \cite{modularity_mucha2010}.
Spectral modularity assigns communities by the trace $\mathrm{Tr}(\mathbf{C}^\top \mathbf{M}\, \mathbf{C})$ of a partition matrix $\mathbf{C}$ against the modularity matrix $\mathbf{M}$ \cite{spectral_modularity_newman2006}; the same bilinear form $\mathbf{C}^\top (\cdot)\, \mathbf{C}$ is exactly the pooling operation that differentiable clustering objectives may optimize end-to-end \cite{dmon_tsitsulin2024}.

A principled partition $\mathbf{C}$ is thus already a pooling assignment, and a scalable, detectability-optimal way to obtain it is a pooling primitive for graphs too large to train on directly.
That these objectives optimize the same block form does not make them interchangeable: differentiable relaxations are sensitive to how the partition is regularized, and a poorly chosen constraint can drive the pooled assignments below graph-theoretic baselines \cite{dmon_tsitsulin2024} --- precisely where such a partition would provide a stable starting point for learning representations.

\smallskip Our work identifies two complementary directions for future research.
One is \textit{neural}: improving differentiable temporal pooling by grounding it in detectability theory --- principled rather than heuristic --- while retaining its block form.
The other is \textit{algorithmic}: attaining the detectability threshold at scale --- through GPU eigensolvers for asymmetric matrices, dispensing with the symmetric reformulation entirely, or through a temporal regularization recovering the same informative directions.
Underlying both is a question we raise for the machine learning and network science communities: when does community structure suffice as a basis for learning on coarse-grained representations, and what more do tasks depending on temporally ordered dependencies rather than co-membership alone require?
Characterizing when community-based pooling preserves these dependencies in the attributed temporal context is, in our view, the open problem this research must strive to answer, and the one toward which it is now directed.

\begingroup
\let\small\footnotesize
\renewcommand{\doi}[1]{}
\renewcommand{\url}[1]{}
\bibliographystyle{splncs04}
\bibliography{references}
\endgroup

\end{document}